\documentclass[runningheads]{llncs}
\usepackage{wrapfig}
\usepackage{graphicx}
\usepackage{multirow}
\usepackage{booktabs}
\usepackage{relsize}
\usepackage{pdflscape}
\usepackage{sidecap}
\usepackage{tikz}

\definecolor{blue}{HTML}{1F77B4}
\definecolor{orange}{HTML}{FF7F0E}
\definecolor{green}{HTML}{2CA02C}

\newcommand{\name}{{\tt ADoPT}}

\begin{document}
\title{ADoPT: LiDAR Spoofing Attack Detection Based on Point-Level Temporal Consistency}
\author{Minkyoung Cho\inst{1} \and
Yulong Cao\inst{2} \and
Zixiang Zhou\inst{1} \and
Z. Morley Mao\inst{1}}
\institute{Computer Science \& Engineering, University of Michigan\\
\email{\{minkycho, zixiangz, zmao\}@umich.edu}\\ \and
NVIDIA Research\\
\email{yulongc@nvidia.com}}
\maketitle              % typeset the header of the contribution
\begin{abstract}
Deep neural networks (DNNs) are increasingly integrated into LiDAR (Light Detection and Ranging)-based perception systems for autonomous vehicles (AVs), requiring robust performance under adversarial conditions. 
One pressing concern is the challenge posed by LiDAR spoofing attacks, where attackers inject fake objects into LiDAR data, leading AVs to misinterpret their surroundings and make faulty decisions. Many current defense algorithms predominantly depend on perception outputs, such as bounding boxes. However, these outputs are intrinsically limited as they are generated by imperfect perception models that process a restricted set of points, acquired from the ego vehicle's specific viewpoint. The reliance on bounding boxes is a manifestation of this fundamental constraint.
To overcome these limitations, we propose a novel framework, named \name~(\underline{\textbf{A}}nomaly \underline{\textbf{D}}etection based \underline{\textbf{o}}n \underline{\textbf{P}}oint-level \underline{\textbf{T}}emporal consistency), which quantitatively measures temporal consistency across consecutive frames and identifies abnormal objects based on the coherency of point clusters. In our evaluation using the nuScenes dataset, our algorithm effectively counters various LiDAR spoofing attacks, achieving a low ($<$ 10\%) false positive ratio and high ($>$ 85\%) true positive ratio, outperforming existing state-of-the-art defense methods, CARLO and 3D-TC2. Moreover, \name~shows promising potential for accurate defense in diverse road environments.

% \keywords{Query optimisation \and Cost Model \and Selectivity Estimation \and Bayesian networks.}
\end{abstract}
\section{Introduction}
\vspace*{-3mm} 
\label{sec:intro}
\looseness=-1
The growing incorporation of deep neural networks (DNNs) in LiDAR (Light Detection and Ranging)-based perception for autonomous vehicles (AVs) calls for rigorous attention to their robust performance. 
In light of this challenge, researchers are focused on developing and refining various defense technologies for potential attacks targeting AV perception systems.
One prominent research direction in this field involves the manipulation of LiDAR point cloud data.
Attackers can fabricate data by jamming and relaying original LiDAR signals~\cite{petit2015remote,yan2016can}, emitting spurious LiDAR purses \cite{shin2017illusion,sun2020towards,sato2022poster,liu2021seeing}, or exploiting vulnerabilities in the DNN-based perception module \cite{yang2021robust,sun2023critical,sun2020adversarial}, causing AVs to misinterpret their driving environment and make faulty decisions (e.g., emergency alarm activation, sudden breaking, lane changing, etc).

\begin{wrapfigure}{r}{0.43\textwidth}
% \vspace{-\baselineskip} 
\centering
    \includegraphics[width=\linewidth]{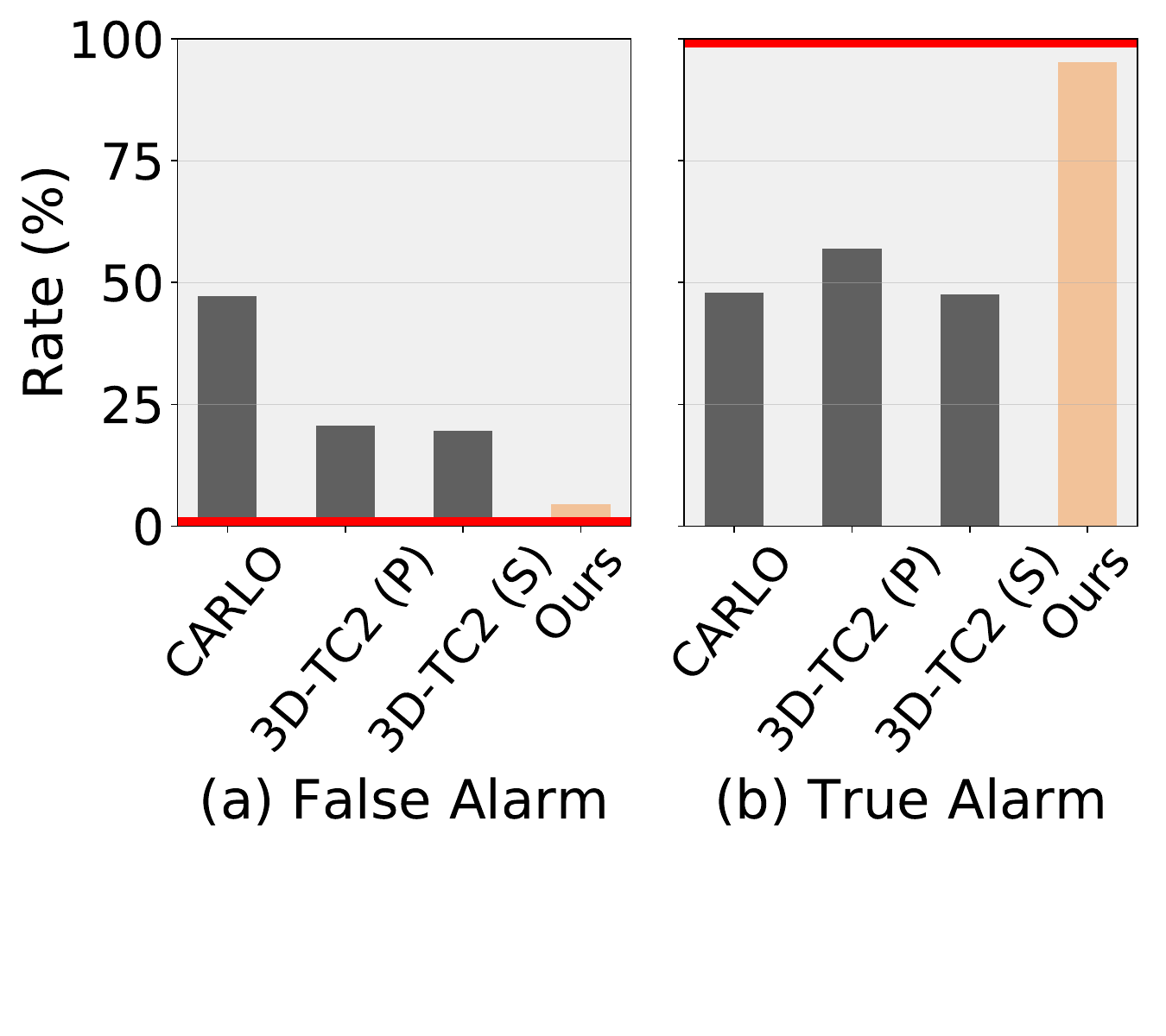}
    % \vspace*{-5mm}     
    \caption{\small{
    Our \name~method outperforms existing methods \cite{you2021temporal,sun2020towards} with a $4.4\sim10.5\times$ lower false alarm rate and $1.7\sim2\times$ higher true alarm rate. 
    The ideal cases are represented by the red solid lines with a false alarm rate of 0 and a true alarm rate of 100.
    We used PointPillars \cite{lang2019pointpillars} for CARLO and 3D-TC2 (P) and SECOND~\cite{yan2018second}  for 3D-TC2 (S) for 3D object detection.
    False alarms arise when benign LiDAR frames are misidentified as attacked, while true alarms occur when poisoned frames, created by injecting simulated pedestrian points, are correctly recognized as attacked. These results show existing defenses often output a bounding box that does not tightly fit the object or is false and struggle to detect small fake objects.}}
    \vspace*{-3mm} 
    \label{fig:motivation}
\end{wrapfigure}

\looseness=-1
Defense algorithms based on AV perception outputs (i.e., bounding boxes) have been widely studied. Figure~\ref{fig:motivation}~showcases two state-of-the-art bounding box-based algorithms: (1) physical principles-based approach (e.g. CARLO)~\cite{sun2020towards,hau2021shadow,xiao2023exorcising}, which detects attacks by leveraging physical principles governing authentic objects; and (2) temporal consistency-based method (e.g. 3D-TC2) \cite{you2021temporal,xiao2019advit,man2023person}, which focuses on motion consistency across adjacent frames.
While temporal consistency offers an edge over physical principles, both are fundamentally limited by their dependence on bounding boxes.
Given the constraints of an ego vehicle's viewpoint and the inherent inaccuracies of perception modules (especially for small and distant objects) \cite{liu2022bevfusion,zhang2021emp,zhang2019accurate}, relying on bounding boxes proves inadequate and leads to inaccuracies in anomaly detection.

We introduce \name~(\underline{\textbf{A}}nomaly \underline{\textbf{D}}etection based \underline{\textbf{o}}n \underline{\textbf{P}}oint-level \underline{\textbf{T}}emporal consistency)~to build perception model-agnostic monitoring modules. Harnessing the rich and comprehensive information present in raw sensor data \cite{zhang2021emp,chen2019cooper}, our approach offers profound advantages in defending against LiDAR spoofing attacks \cite{sun2020towards,you2021temporal,xiao2023exorcising}, bypassing the limitations of traditional perception models.
While raw sensor data provides extensive information, implementing defense algorithms (e.g. temporal consistency) based on raw sensor data is challenging.
Our approach emerges from our observation of the intuitive notion that an object consists of point clusters with a specific degree of point intensity, moving coherently.
This understanding enables the measurement of temporal consistency at the point cloud level.
Utilizing our temporal consistency foundation, we present a two-stage approach to detect adversarial manipulations in frames. Initially, our coherence-enhanced scene flow estimation predicts expected object locations while maintaining coherency, even in the face of point injections, outperforming conventional methods susceptible to anomalies. 
This robust estimation sets a firm foundation for the subsequent phase: clustering-based anomaly detection. It contrasts point clusters between expected and observed point locations, with discrepancies between them highlighting potential adversarial interventions.

\name~stands robust against both dense and sparse point injection LiDAR spoofing attacks, achieving a commendable false positive ratio (FPR) of less than 10\% and a true positive ratio (TPR) exceeding 85\%, thereby surpassing existing defense mechanisms grounded on perception output. 
Additionally, we highlight the effectiveness of our anomaly detection metric based on cluster coherency, showing its superiority over traditional methods through comparative analysis.

\vspace*{-3mm} 
\section{Related Work}
\vspace*{-3mm} 
Autonomous vehicle (AV) defense mechanisms against sensor data fabrication attacks are broadly classified into physical principle-based and consistency-based methods.

\noindent\textbf{Physical Principle-based Defense.}
These techniques leverage specific geometries or physical-invariant properties, which attackers struggle to imitate when forging objects. CARLO~\cite{sun2020towards} employs free/occluded space or laser rays within a frustum space related to each detected bounding box to distinguish between real and fake objects. Shadow-Catcher~\cite{hau2021shadow} utilizes shadow region differences based on bounding box coordinates, and LOP \cite{xiao2023exorcising} introduces the concept of \textit{objectness} by considering the point density and the distance from the LiDAR sensor to the predicted objects. These approaches use specialized rules to assess an object's adherence to physical principles.

\noindent\textbf{Consistency-based Defense.}
These methods emphasize temporal consistency and show promising detection success rates for AV systems. They exploit the invariant nature of object motion across consecutive frames. AdvIT \cite{xiao2019advit} counters adversarial attacks on video frames, where the attacker manipulates the distribution of points but preserves the appearance of the original points, by estimating the optical flow of each pixel and measuring temporal consistency. PercepGuard \cite{man2023person} utilizes spatio-temporal consistency for misclassification attacks, where the attacker alters the labels of detected outputs on camera images (e.g., from car to people), and verifies moving patterns of bounding boxes. 3D-TC2 \cite{you2021temporal} proposes a temporal consistency check-based method to detect LiDAR spoofing attacks, converting LiDAR point clouds into 2D images and comparing predicted motion to detected bounding boxes.

\noindent\textbf{Limitations.}
Existing studies rely on perception modules, assuming their high accuracy. However, raw sensor data processing using perception modules can result in false detections or information loss, especially for small objects like pedestrians and cyclists -- critical objects that autonomous vehicles must consider in making navigation decisions. 
Our work introduces a novel paradigm for attack detection algorithms, using only raw sensor data to achieve robust defense regardless of the object type. We detail the core components enabling point-level anomaly detection in the following sections.

\vspace*{-3mm} 
\section{Background: Scene Flow Estimation}
\vspace*{-3mm} 
In the 3D point cloud domain, scene flow represents the 3D motion of each point across consecutive frames. Accurate scene flow estimation is crucial for predicting user or AV motion and estimating trajectories. However, real-world estimation remains challenging due to temporal occlusion and dynamic, rigid object motions. Scene flow estimation has evolved into two primary branches.

\noindent\textbf{Offline Learning Methods.} These approaches \cite{liu2019flownet3d} use separate offline training processes with annotated datasets. Scene flow estimation is formulated as a DNN model that receives a pair of frames and outputs the optimal flow. DNN models offer customizability and high capacity for flow representation, achieving high accuracy while addressing bad correspondences. However, they face limitations such as requiring substantial data and ground truth labels, which are difficult to obtain \cite{chang2019argoverse,caesar2020nuscenes}. Researchers generate labels using alternative methods \cite{pontes2020scene} or employ self-supervised learning \cite{wu2020pointpwc,mittal2020just}. These methods may struggle with input frames deviating from the training dataset due to low generalization capabilities~\cite{li2021neural}.

\noindent\textbf{Online Optimization Methods.} These approaches do not require separate training processes or datasets, instead formulating scene flow estimation as an optimization problem. 
% They demonstrate higher accuracy on out-of-distribution point cloud frames compared to offline learning-based methods, making them more suitable for real-world situations.
They demonstrate higher accuracy on out-of-distribution point cloud frames, which fall outside the training dataset used by offline learning-based methods, making them more suitable for real-world situations.

Classic methods, such as Iterative Closest Point (ICP) \cite{besl1992method}, initialize flow vectors for the point within the point cloud frame, solving optimization problems to find the optimal flow vectors that represent the discrepancy of two input frames at runtime.
Recent studies have proposed various solutions with high generalizability by formulating scene flow estimation in diverse ways, such as multi-layer perceptron (MLP) \cite{li2021neural,li2022non}, graph Laplacian \cite{pontes2020scene}, or Bayesian inference \cite{hirose2020bayesian}. 
Among these methodologies, NSFP \cite{li2021neural} introduced an advanced method by changing flow representation from displacement vectors to an MLP-based model. With this MLP-formed flow, they solve the optimization problem and find the optimal flow at runtime while iteratively adjusting the MLP parameters like DNN training.

\begin{figure}[tp]
  \begin{minipage}[t]{0.24\linewidth}
    \centering
    \includegraphics[width=\linewidth]{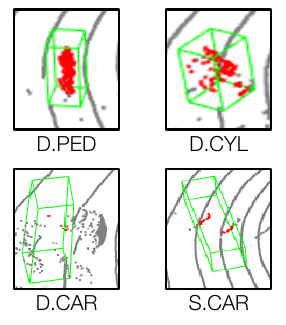}
  \end{minipage}
  \hfill
  \begin{minipage}[t]{0.75\linewidth}
  \vspace*{-33mm} 
    \caption{Examples of Injected Points. D.PED, D.CYL, and D.CAR are fake objects created by dense point injection attacks, mimicking a pedestrian, cyclist, and car, respectively. S.CAR results from a sparse point injection attack imitating a car. AVs recognize these objects as real (green bounding box). Red points are within the bounding box, while gray points are outside, suggesting that bounding boxes may not always adequately fit objects.}
    \vspace*{3mm} 
    \label{fig:attacks}
  \end{minipage}
\end{figure}

\vspace*{-3mm} 
\section {Threat Model}
\vspace*{-3mm} 
We investigate two spoofing attacks: dense and sparse point injection. Dense point injection attacks inject up to 200 points and achieve a high Attack Success Rate (ASR) of 96\%-97\%, producing a visually recognizable fake object. 3D-TC2 \cite{you2021temporal} is designed to counter this attack and serves as our evaluation baseline. Conversely, sparse point injection attacks \cite{sun2020towards} inject up to 64 points, rendering the fake object difficult to visually identify, with an ASR of less than 21\%. CARLO \cite{sun2020towards} is a proposed defense method to combat this sparse injection attack, used as our evaluation baseline. Figure~\ref{fig:attacks} displays examples of spoofed objects. 

\vspace*{-3mm} 
\section{\name~Methodology}
\vspace*{-3mm} 

\begin{figure}[tp]
    \centering
    % \vspace*{-2mm} 
    \includegraphics[width=1\linewidth]{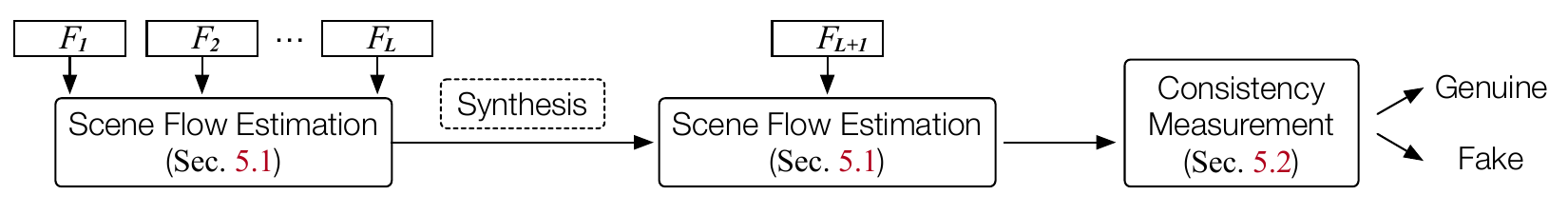} 
  \vspace*{-3mm} 
   \caption{\name~Architecture.}
    \vspace*{3mm} 
   \label{fig:mechanism}  
\end{figure}

% Our primary objective is to eliminate the high dependence on imperfect perception outputs on ego vehicle's limited viewpoint by realizing the point-level anomaly detection, thus reducing the false detection and enhancing the true detection. 
\looseness=-1
In this section, we introduce \name, a solution for point-level anomaly detection devised to enhance the resilience of object detection systems. 
Leveraging the observation that injected points demonstrate \textit{poor temporal consistency} — appearing inconsistently within the point cloud frame over time — \name~utilizes scene flow estimation to quantify objects' temporal consistency, thereby facilitating the detection of point injection attacks.
Figure~\ref{fig:mechanism} illustrates overall \name~architecture, where $F_1, F_2, ..., F_L$ are sequential historical LiDAR point cloud frames, and $F_{L+1}$ is the subsequent incoming frame.
Initially, \name~generates a synthesis of preceding frames by aligning all points from the historical frames using scene flow estimation (Sec.~\ref{sec:coherence_sfe}). This synthesized representation is then compared with the incoming frame to identify points that showcase inadequate temporal alignment, earmarking them as potential injections from attackers (Sec.~\ref{sec:comp_consistency}).
Furthermore, we outline several techniques to mitigate runtime overhead, thereby making \name~a viable solution when implemented in AV systems (Sec.~\ref{sec:other_comp}).

\vspace*{-4mm} 
\subsection {Coherence-Enhanced Scene Flow Estimation}
\vspace*{-3mm} 
\label{sec:coherence_sfe}
For quantification of temporal consistency, central to \name\ is the process of aligning points captured at different timestamps and combining them into a single frame. 
Scene flow estimation (SFE) is crucial for aligning point cloud frames by calculating optimal point displacement \cite{besl1992method,li2020evaluation}. For generalizability to various road environments and different injected objects, we employ an MLP-formed neural prior to represent scene flow and optimize the MLP parameters at runtime for a pair of point cloud frames ($F_1$, $F_2$), inspired by NSFP~\cite{li2021neural}. 

\begin{wrapfigure}{r}{0.43\textwidth}
    \vspace*{2mm} 
    % \vspace{-\baselineskip} 
    \centering
    \includegraphics[width=\linewidth]{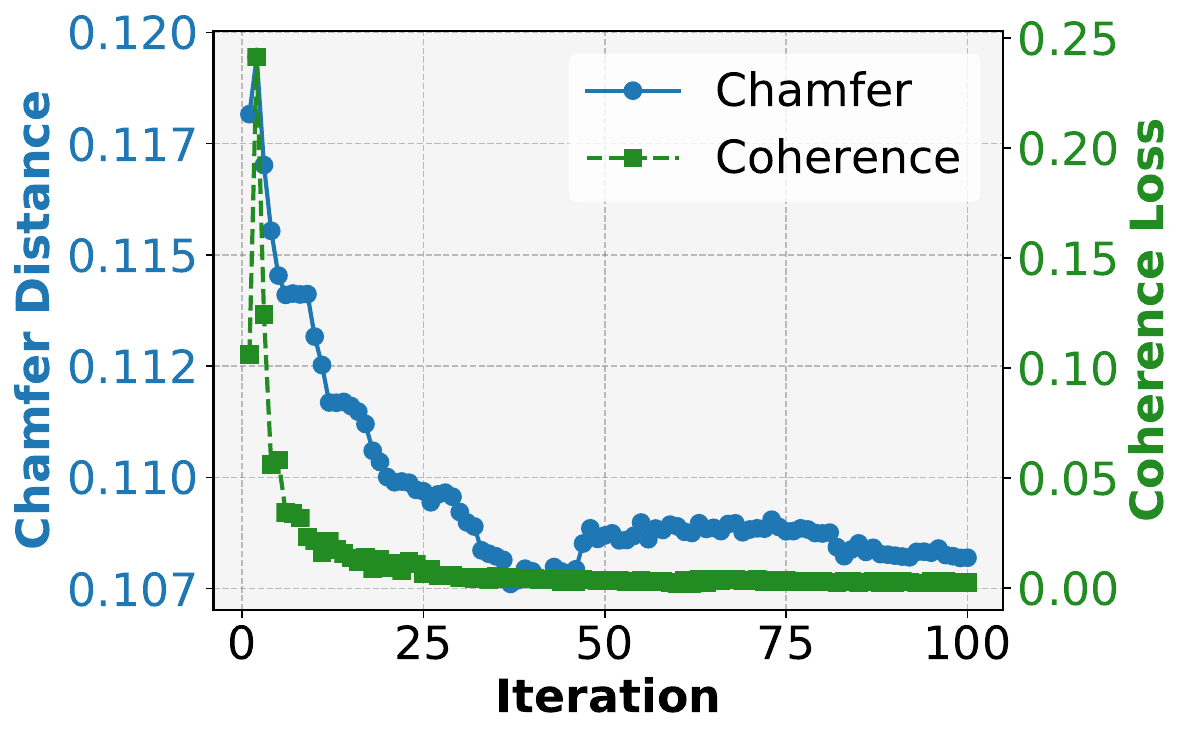}
    % \vspace*{-2mm} 
    \caption{As the number of iterations increases, the two loss values converge harmoniously without impeding each other's individual convergence.}
    \vspace*{2mm} 
    \label{fig:losschange}
\end{wrapfigure}

SFE serves a dual purpose: it aligns points in historical frames and juxtaposes the resulting synthesized frame with newly arriving one.
However, ensuring precise scene flow becomes challenging without confirmed temporal consistency. 
In the presence of a LiDAR spoofing attack, conventional methods frequently falter, predicated on the assumption of consistent object appearances across frames. 
This vulnerability arises because this assumption can be violated through the continued use of SFE over a series of frames, or through the introduction of spurious objects. 
Repeatedly deploying SFE may amplify errors, leading to a dispersion effect in the synthesis. 
Additionally, the presence of fabricated points in recent frames hampers the precise functioning of SFE, creating erroneous point correspondences between synthesized and fake points.

To counteract these challenges, we have conceptualized temporal consistency by viewing objects as cohesive point clusters with inherent intensity. 
Recognizing this coherence, we introduce a loss function that enhances the coherence between the motion flows of points belonging to each cluster (i.e., part of an object). Here, we use a clustering method, DBSCAN~\cite{ester1996density}, a well-established spatial clustering algorithm, to verify if two given points are part of the same cluster. By defining the following loss term, we can enforce neighboring points to move coherently:
% \vspace*{-1mm} 
\begin{center}
${L}_{coherence}(F_1)=\frac{1}{N^2} \mathlarger{\sum}\limits_{p_i,p_j \in F_1} (M(p_i, p_j) \cdot w(p_i, p_j) \cdot ||fl(p_i)-fl(p_j)||^2)$
\end{center}
% \vspace*{-2mm}
\noindent
where $M(p_i, p_j)$ is a binary clustering mask indicating whether any two points in $F_1$, $p_i$ and $p_j$, belong to the same cluster. Concurrently, $w(p_i, p_j)$ is the weight value between these points, formulated to foster more coherent movement between closer points by being influenced by the distance between them.
$fl(p_i)$  is a 3-dimensional flow vector that indicates the displacement of point $p_i$ along the x, y, and z axes.
$N$ denotes the number of points included in valid clusters (i.e., non-outlier points) as identified through DBSCAN.

Consequently, our final loss function is defined by a combination of the Chamfer Distance (CD) \cite{fan2017point} and coherence loss, enabling us to find the optimal scene flow that represents the point motions between $F_1$ and $F_2$ while preventing any point from deviating from its original cluster.
% \vspace*{-3mm} 
\begin{center}
${L} = \alpha {L}_{chamfer}(F_1, F_2) + \beta {L}_{coherence}(F_1)$
\end{center}
\vspace*{1mm} 
\noindent
Here, ${L}_{chamfer}(F_1, F_2)$ is the CD value, the most popular distance metric for two point cloud sets, which is defined as:

% \vspace*{-5mm}
\begin{center}
${L}_{chamfer}(F_1, F_2) = \mathlarger{\sum}\limits_{p_i \in F_1}\min\limits_{q_j \in F_2}||p_i-q_j||_2^2 + \mathlarger{\sum}\limits_{q_j \in F_2}\min\limits_{p_i \in F_1}||p_i-q_j||_2^2$
\end{center}
% \vspace*{-3mm}
\noindent where $p_i$ and $q_j$ represent individual points in $F_1$ and $F_2$ respectively. 
Figure~\ref{fig:losschange} showcases the functioning of our proposed loss function during the online optimization process, emphasizing the harmonious convergence of the two loss terms.

\vspace*{-3mm} 
\subsection {Cluster-based Consistency Measurement}
\vspace*{-2mm} 
\label{sec:comp_consistency}
Implementing the proposed SFE method produces a warped synthesis by predicting the appearance of individual points at the moment each incoming frame arrives. A pivotal stage in \name~is the temporal consistency measurement between the resulting synthesis and the incoming frame. 
However, results obtained using conventional distance metrics (e.g., Chamfer Distance), often lead to substantial variances, thereby making the discrimination process markedly challenging.
This fluctuation is predominantly driven by the differing number of points in individual frames, a variable greatly affected by the intricacies of various road scenarios. Moreover, the unpredictable changes introduced by attackers, affecting both the location and the number of points, deem traditional handcrafted schemes inadequate. This necessitates a suitably designed metric. 

Our approach employs a cluster-based metric based on the understanding that objects naturally form groups of point clusters. Recognizing the potential in the overlapping characteristic of the synthesis and incoming frame provides a strategy for assessing their consistency. To foster this strategy, we first meld the warped synthesis with the incoming frame, distinctly marking each point to denote whether it originates from the synthesis or the incoming frame. 
This meticulous categorization aids in identifying areas of inadequate temporal alignment, enabling reliable detection of fabricated objects.
Subsequently, \name~uses DBSCAN to identify local clusters, effectively discarding outliers, and then removes clusters that contain synthesis points, a process which verifies the presence of genuine objects consistent with historical objects. In benign scenarios, this tactic results in a complete absence of clusters, a testament to the adequate temporal alignment across frames. 
However, in poisoned scenarios, clusters exclusively formed of incoming frame points remain, serving as indicators of fabricated objects. Illustrative examples are shown in Appendix~\ref{supp_consistencymetric}. 

\vspace*{-3mm} 
\subsection {Additional Components for Runtime Overhead Reduction}
\vspace*{-1mm} 
\label{sec:other_comp}
\label{sec:key_components}
\begin{figure}[pt]
    \begin{center}
    % \vspace*{-1mm}
     \includegraphics[width=1\linewidth]{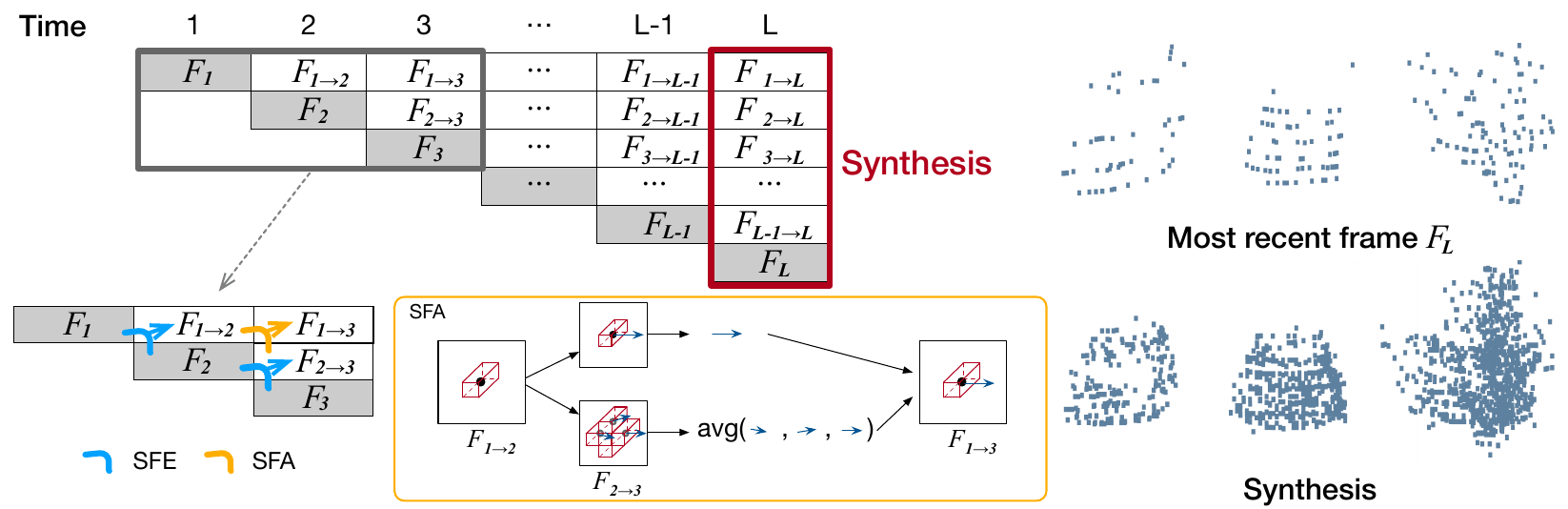} 
    \end{center}
    % \vspace*{-6mm} 
    \caption{Synthesis Generation Procedure. Given historical frame length $L$, a new LiDAR frame enters the system for Scene Flow Estimation (SFE). The estimated flow propagates through Scene Flow Approximation (SFA) to upper cells. The top right figure shows the actual input frame $F_L$, and the bottom right displays the generated synthesis with clear and dense object shapes (three cars, the rightmost one is obscured behind a tree).}
   \label{fig:sf_approx}
    \vspace*{2mm} 
\end{figure}
\noindent \textbf{Synthesis Generation on Historical Frames.}
To achieve our goal of identifying anomalous objects, \name~needs to differentiate between suddenly appearing normal objects and maliciously placed objects. \name~synthesizes past frames by warping them to the time of the last historical frame $F_L$ using SFE, inspired by point cloud densification techniques \cite{wang2022neural,li2021neural}. However, these methods are computationally intensive, and generating synthesis may take up to $O(NL^2)$ time, assuming latency in solving optimization problems is $N$, and we use $L$ historical frames. For instance, with a 0.1-second latency and 10 historical frames, it would take 10 seconds to process all the scene flow estimations, which is impractical and cannot be concealed through parallelization.

To reduce computational complexity, \name~approximates scene flow instead of solving optimization problems. 
% We propagate the estimated scene flow between a specific past frame and its following frame, allowing us to estimate scene flow once at each timestamp.
The estimated scene flow is propagated to its preceding frames, facilitating a single estimation of the scene flow at each timestamp.
% As LiDAR senses a varying number of points, we use voxelization to identify corresponding points to propagate the most recent scene flow into past frames. We map voxel indices between frames to add the corresponding voxel's scene flow to the target point. If there are no mapped points, we compute the target voxel's scene flow by averaging the adjacent voxels' scene flows, as shown in Figure~\ref{fig:sf_approx}. 
As LiDAR senses a varying number of points, voxelization is utilized to identify corresponding points in order to propagate the most recent scene flow into past frames. This process involves mapping voxel indices between frames to integrate the corresponding voxel's scene flow with the target point. In instances where there are no mapped points, the target voxel's scene flow is computed by taking the average of the adjacent voxels' scene flows, a strategy depicted in Figure~\ref{fig:sf_approx}.
This substantially reduces the time complexity of computation-intensive SFE processes from $L^2$ to $L$, 
saving time and enabling parallelization with the SFE process.

\noindent \textbf{Voxel Downsampling.}
In \name, the total latency is primarily influenced by the number of points. To optimize the system, minimizing latency without significant TPR loss and FPR gain, we reduce the number of points via voxelization. This approach's efficacy is detailed in Sec.~\ref{sec:experiments}, where we illustrate the interplay between voxel grid size, total latency, and the accuracy of attack detection. 
This highlights the critical role of selecting the optimal voxel grid size to balance system accuracy with timely execution.
\vspace*{-3mm}
\section{Evaluation}
\vspace*{-2mm}
In this section, we evaluate \name~under different attack scenarios. Additionally, we compare our proposed algorithm with two widely-used defense methods for LiDAR spoofing attacks, CARLO and 3D-TC2. All experiments are conducted on a server equipped with two Intel Xeon 4110 CPUs and one NVIDIA RTX 2080 GPU.

\noindent \textbf{Dataset.}
Our evaluations are performed on the nuScenes dataset~\cite{caesar2020nuscenes}, a large-scale autonomous driving dataset collected from vehicles equipped with a 32-beam LiDAR system. The nuScenes dataset is divided into two subsets: v1.0-mini (comprising 10 scenes) and v1.0-trainval (comprising 350 scenes). Each scene is 20 seconds long and annotated at a frequency of 2 Hz.

\noindent \textbf{Attack Scenarios.}
The poisoned dataset utilized in the dense point injection attack is sourced from the authors of 3D-TC2, who leveraged the v1.0-mini dataset. 
In this attack, spoofed data points that represent vehicles, cyclists, and pedestrians are introduced systematically. 
Concurrently, in addressing the sparse point injection attack, we generated 355 poisoned frames using the validation set of the v1.0-trainval dataset, which consists of 150 scenes, adhering to the approach presented in CARLO. 
Given the intrinsic sparsity characteristic of this attack, only spoofed points representative of vehicles are introduced.

\noindent \textbf{Parameter Setting.}
We opt for a multi-layer perceptron (MLP) architecture composed of six layers and 128 hidden units, a configuration empirically determined to yield the highest accuracy while maintaining a low latency on our dataset.
Figure \ref{fig:losschange} shows loss convergence after 30 iterations, influencing our choice of a 30-iteration count. 
For training, we use a fixed learning rate of 0.0008, empirically derived for optimum performance and convergence potential.
In defining our loss function, we attribute values of 1 and 2 to variables $\alpha$ and $\beta$, respectively, and the weights between the points, $w(p_i, p_j) $, are all set equally to 1.
The DBSCAN procedure relies heavily on two critical thresholds: the minimal distance between the nearest points and the minimal count of points necessary to form a valid cluster. 
Carefully optimizing for FPR and TPR, we established thresholds of at least 17 points and a 0.25 distance parameter for dense point attacks, and a minimum of 9 points with a 0.75 distance threshold for sparse point attacks.
Detailed insights into the threshold determination process are elaborated in Appendix~\ref{supp_thresholdset}.

\vspace*{-3mm}
\subsection{Experimental Results}
\label{sec:experiments}
\vspace*{-1mm}
\looseness=-1

\begin{table}[tp]
    \centering
    \caption{Comparison of Defense Methods. A lower false positive rate (FP) and a higher true positive rate (TP) indicate a more accurate attack detection.}
    \vspace*{2mm} 
    \label{tab:eval_different_objects}
    \scalebox{0.85}{
    \begin{tabular}{lcccccc} 
        \toprule
        \multirow{2}{*}{} & \multicolumn{4}{c}{Dense Point Injection} & \multicolumn{2}{c}{Sparse Point Injection} \\
        \cmidrule(lr){2-5} \cmidrule(lr){6-7}
        & FP $\downarrow$ & TP (D.CAR) $\uparrow$ & TP (D.CYL) $\uparrow$ & TP (D.PED) $\uparrow$ & FP $\downarrow$ & TP (S.CAR) $\uparrow$ \\
        \midrule
        CARLO \cite{sun2020towards}      & 47.2  &    48.0     & 49.4        & 48.0        & 47.9     & 54.4 \\
        3D-TC2 (PP) \cite{you2021temporal}   & 20.7 & 98.6      & 95.0     &    56.9    & 16.6    & 53.5 \\
        3D-TC2 (SEC) \cite{you2021temporal}  & 19.6  & 98.3      & 45.8       &  47.5     & 16.3     & 84.2 \\
        % \name^{\textstyle\ast}         & 8.0  & 96.3       & 95.7        & 94.9       & 9.9     & 85.9 \\
        $\name$ & 4.5  & 97.2       & 98.3        & 95.2      & 9.3     & 85.4 \\
        \bottomrule
    \end{tabular}
    }
    \vspace*{2mm} 
\end{table}

\begin{figure}[t]
\centering
% \vspace*{-2mm}
\includegraphics[width=1\linewidth]{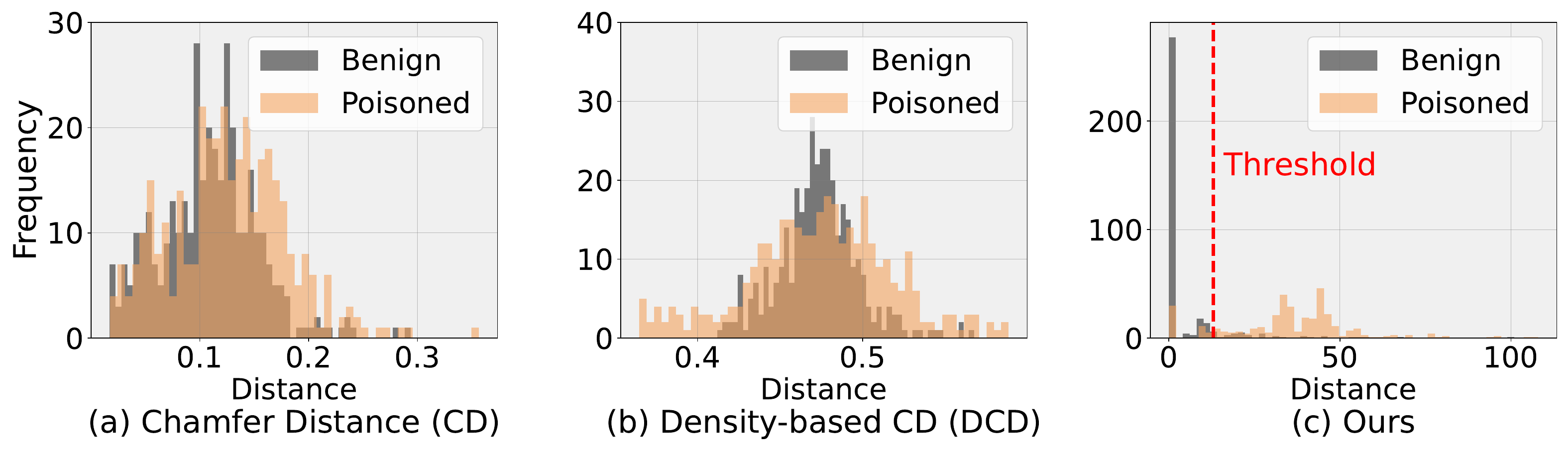}
% \vspace*{-7mm}
\caption{Comparison of Anomaly Detection Metrics: (a) Chamfer Distance, (b) Density-based Chamfer Distance, and (c) Our Proposed Cluster-based Metric. The x-axis represents the distance values, and the y-axis indicates the number of frames corresponding to each value. Our approach allows for establishing a threshold for attack detection, unlike conventional metrics. The difference between the average distance values for benign and poisoned cases supports this claim: (a) benign: 0.11, poisoned: 0.13, (b) benign: 0.47, poisoned: 0.47, (c) benign: 4.31, poisoned: 35.34. In this work, we set our distance threshold to 15.}
\label{fig:design_choice_q2}
\vspace*{3mm} 
\end{figure}

\noindent \textbf {Effect of \name.}
To substantiate the effectiveness of \name, we benchmark its performance against baseline methods under dense and sparse point injection attack scenarios (see Table~\ref{tab:eval_different_objects}). 
In this evaluation, we focus on two pivotal metrics: the false positive rate (FPR), denoting the incorrect identification of benign frames as attacked, and the true positive rate (TPR), reflecting the correct detection of poisoned frames.
Our method manifests a low FPR, evincing its efficacy in curtailing false alarms during the detection of spoofing attacks. 
We match the baseline accuracy on relatively large objects such as D.CAR, while substantially exceeding it when it comes to smaller objects, notably achieving TPRs of 98.3\% and 95.2\% for cyclists and pedestrians, respectively. 
These statistics underline \name's superior ability to pinpoint small spoofed objects. 
Consequently, \name~demonstrates marked supremacy in identifying LiDAR spoofing attacks across different object types and attack scenarios.

\begin{wrapfigure}{r}{0.6\textwidth}
    \centering
    % \vspace*{-5mm} 
    \includegraphics[width=\linewidth]{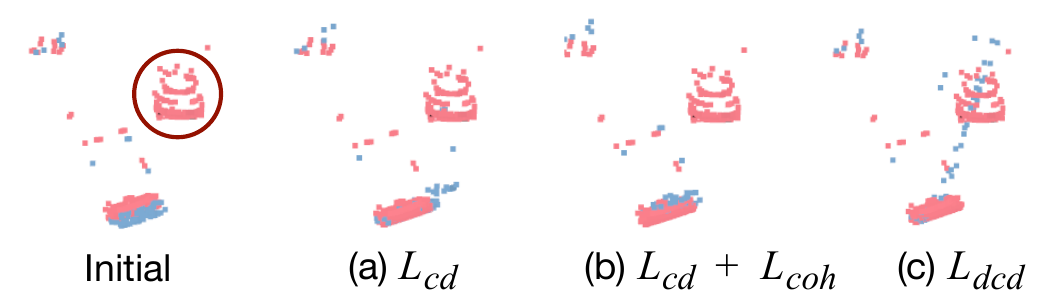}
    % \vspace*{-6mm} 
    \caption{Ablations with/without Coherence Loss. We examine the alignment of two point cloud sets (red and blue points) by comparing the warped blue points (obtained by adding the estimated scene flow) to the red points. 
    The first image denotes their initial status, and the car (circled in red) indicates a fake object.
 (a)~shows results using only CD without coherence loss; (b) displays outcomes of our proposed method using both CD and coherence loss. We also explore replacing our loss with DCD in (d).
This comparison reveals that fake points hinder SFE, seemingly drawing the red points towards them, highlighting the necessity of coherence-enhancing loss under adversarial settings.} 
% This comparison shows the fake points disrupt SFE by attracting points toward themselves, emphasizing the importance of our coherence-enhancing loss under adversarial conditions.}
    % \vspace*{3mm} 
    \label{fig:ablation}
\end{wrapfigure}

\noindent \textbf {Clustering-based Metric for Anomaly Detection.}
In considering alternative design approaches for consistency measurement, we acknowledge the potential utility of established distance metrics. 
Among the prevalent metrics for evaluating point cloud similarity are CD and Earth Mover's Distance (EMD) \cite{fan2017point}. Despite its utility, EMD's computational demands deem it unfit for attack detection. 
To this end, our analysis leverages the Density-based Chamfer Distance (DCD) \cite{wu2021density}, a novel metric that synergizes the strengths of CD and EMD, promising enhanced accuracy.

\begin{wrapfigure}{r}{0.33\textwidth}
% \vspace*{-10mm}    
\hspace{1mm}
\centering
    \includegraphics[width=\linewidth]{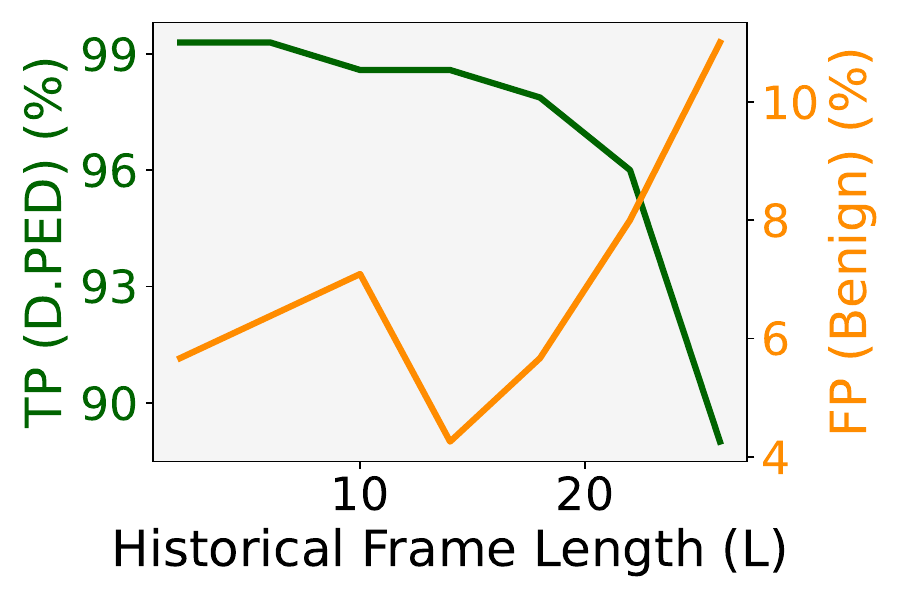}
    % \vspace*{-3mm}    
     \caption{Attack Detection Accuracy Changes in Relation to Historical Length}
% \vspace*{2mm}     
\label{fig:hist_length}
\end{wrapfigure}

As illustrated in Figure \ref{fig:design_choice_q2}, both CD and DCD exhibit substantial fluctuations according to road configurations, with object count and road environment serving as prominent influencing factors. 
This variability creates a challenging environment for establishing the ideal thresholds to distinguish between benign and poisoned frames.
In contrast, our clustering-based metric enables threshold determination for attack decisions, making it the most reasonable metric for our situations.

\noindent \textbf{Ablation Study.}
We conducted an ablation study to assess the impact of including or excluding the coherence loss term, as well as employing DCD. 
Further details can be found in Figure~\ref{fig:ablation}.
Utilizing DCD as our loss function in dense point injection attack scenarios yielded a 4.6\% FPR and TPRs of 94.8\%, 92.9\%, and 94.0\% for D.CAR, D.CYL, and D.PED, respectively. These findings, exhibiting stable FPR and decreased TPRs, support the idea that the injected fake points compromise accurate SFE, emphasizing the effectiveness of the coherence-enhanced SFE.

\noindent\textbf{Impact of Historical Frame Length.} Historical frames help distinguish sudden appearances of valid and fake points by correlating them with previously observed points. We utilize 10 historical frames at a 10 Hz frequency to align with 3D-TC, allowing for a fair comparison of the effects of using raw data vs. using detection output. Testing with various frame lengths produces nearly consistent results from lengths of 2 to 15, but shows a decline thereafter (Figure~\ref{fig:hist_length}).

\vspace*{-3mm}
\subsection{Limitation \& Analysis}
\label{sec:limitation}
\begin{wrapfigure}{r}{0.4\textwidth}
    \centering
    \vspace{2mm} 
    \includegraphics[width=1\linewidth]{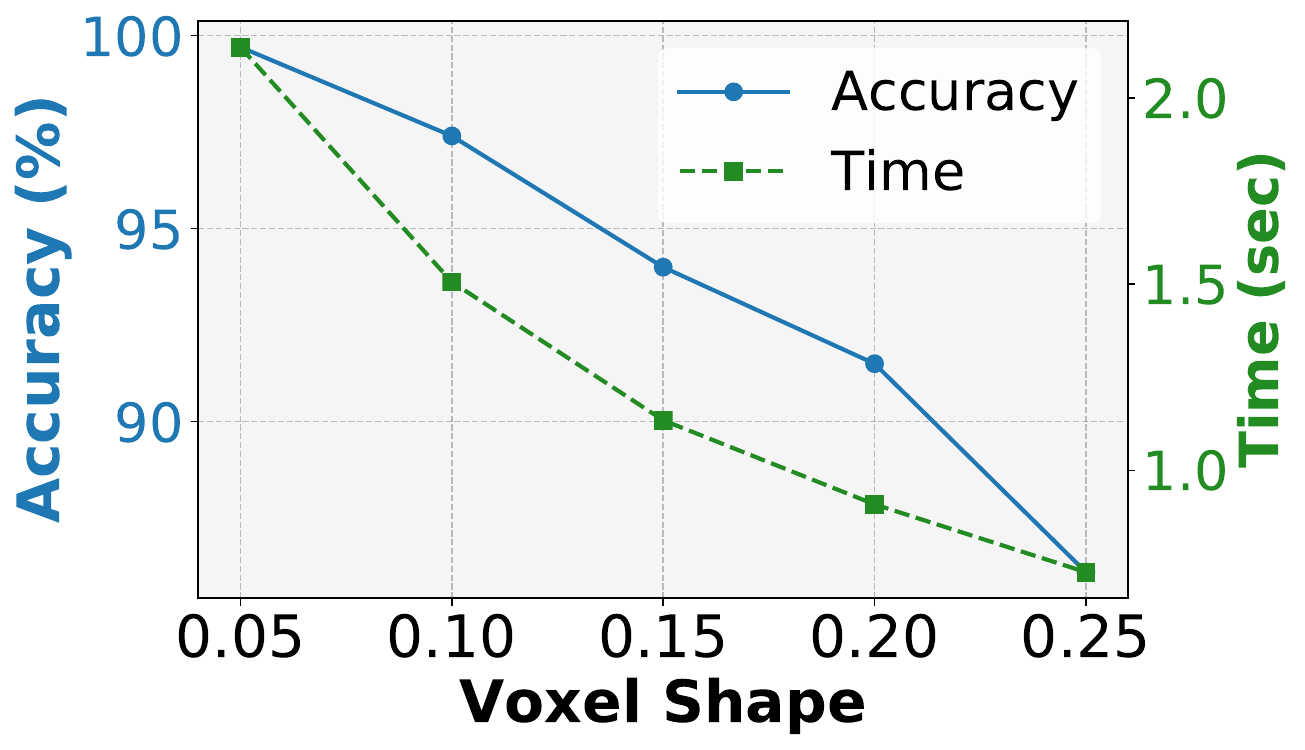}
    % \vspace*{-6mm} 
    \caption{Impact of Voxel Shape on Attack Detection Accuracy and Time. We use a cubic-shaped voxel, and the value of the x-axis represents the length of one side of the voxel.}
    \vspace*{2mm} 
    \label{fig:overhead}
\end{wrapfigure}

\vspace*{-1mm}
\noindent\textbf{Runtime Overhead.} In \name, the primary contributor to total latency is SFE latency, significantly influenced by the voxel shape. As illustrated in Figure~\ref{fig:overhead}, \name~currently experiences a considerable latency of 2.1 seconds. Nonetheless, we have the capacity to diminish this to a mere 0.7 seconds, sustaining a robust attack detection accuracy at roughly 87\% of the initial accuracy. This indicates a marked enhancement in speed without a critical sacrifice in detection capability. To further hasten this process, we envision incorporating mixed-precision training techniques \cite{nvidiasoc}, utilizing a 16-bit (or lower) numerical format for MLP parameters. We also aim to reduce runtime overhead through the pre-training of the MLP model, coupled with the application of test-time training \cite{sun2019test}, a strategy for quick convergence and adaptability to various incoming frames.

\noindent \textbf{Failure Cases}. 
As we employ the spatial clustering method for attack detection, most failure cases arise when spoofed objects are attached to benign road objects. Although classified as false negatives, the spoofed object is identified as part of the benign object it is attached to; thus, it does not significantly affect existing navigation decisions or trigger numerous sudden alarms. Refer to Appendix~\ref{supp_failure} for point cloud data illustrating a failure situation.

\noindent\textbf{Analysis on Impact per Component on Performance.}
\name's performance is contingent upon the accuracy of both SFE and DBSCAN. However, thanks to its structural advantage, it is able to maintain effective responsiveness even when there is a decrease in the accuracy of either of these elements. Let's explore this further:
Firstly, low SFE accuracy can lead to flawed synthesis and clustering owing to inaccuracies in frame warping, thereby generating false positive errors. This issue is often caused by shifts in input data distribution and occlusion. However, our coherence-enhanced SFE, grounded in an online optimization solving detailed in Section~\ref{sec:coherence_sfe}, effectively counters this shift issue, helping to avoid low SFE accuracy and associated false positive errors.
Secondly, despite adequate SFE, poor clustering accuracy can occur when real objects come too close to each other or when fake objects attach to real ones, resulting in merged object clusters and either false positive or negative errors. To mitigate this, we leverage extended historical frames to foster a broader understanding of the context, which aids in discerning the real objects from the fake ones more accurately. It is important to note that false negatives generally do not significantly impair navigation or trigger false alarms since a real object is indeed present, as discussed in the Failure Cases section.
Lastly, when both SFE and clustering accuracy are satisfactory, \name efficiently identifies attacks in the majority of cases, as evidenced by the data presented in Table~\ref{tab:eval_different_objects}.

On the other hand, while \name~operates independently of LiDAR object detection, immediate elimination of fake objects can enhance AV perception accuracy by reducing falsely detected instances.

\vspace*{-3mm}
\section{Conclusion}
\vspace*{-3mm} 
We present the \name~framework, designed to detect LiDAR spoofing attacks on AVs by measuring temporal consistency at the point cloud level.
\name~surpasses existing methods, delivering lower false positive rates and higher true positive rates. 
% Notably, \name~adapts to diverse driving scenarios due to its runtime optimization feature. 
While currently focused on single-frame fake object injection attacks, \name~has the potential to address LiDAR spoofing attacks spanning consecutive frames by promptly eliminating detected spoofed points, converting them into benign frames, and continuously identifying spoofed objects in subsequent incoming frames.
In the future, we aim to evolve \name~to counter diverse LiDAR data manipulation attacks (e.g., object removal attacks), thereby enhancing the robustness of perception modules in AVs.

\vspace*{-3mm}
\section{Acknowledgements}
\vspace*{-3mm}
We express our gratitude to Jiachen Sun and Qingzhao Zhang for their discussions on this work, and to our area chairs and anonymous reviewers for their invaluable feedback.
This work is partially supported by NSF under grants CNS-1930041, CMMI-2038215, and Grant \# 2112562 (the National AI Institute for Edge Computing Leveraging Next Generation Wireless Networks). 

\newpage
\bibliographystyle{splncs04}
\bibliography{reference}

\newpage
\vspace*{-10mm} 
\section*{Appendix}
\appendix

\section{Cluster-based Consistency Measurement}
\label{supp_consistencymetric}
\begin{figure}[htp]
  \centering
  % \vspace*{-5mm}
  \begin{minipage}[b]{0.65\textwidth}
    \includegraphics[width=1\linewidth]{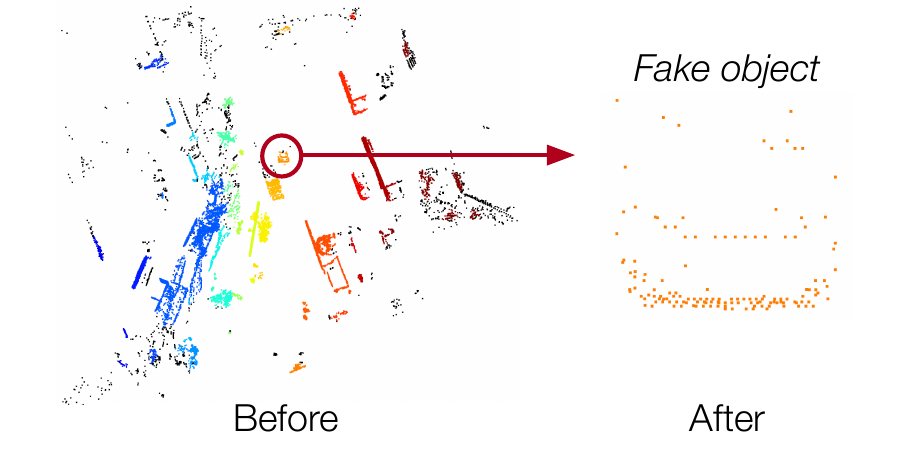}
    \caption{Qualitative Results from Spatial Clustering. Using predetermined thresholds, we identified 53 distinct clusters, each represented by a unique color, with black indicating outliers.}
    \label{fig:2}
  \end{minipage}
  \hspace{0.01\textwidth}
  \begin{minipage}[b]{0.3\textwidth}
    \centering
    \includegraphics[width=0.9\linewidth]{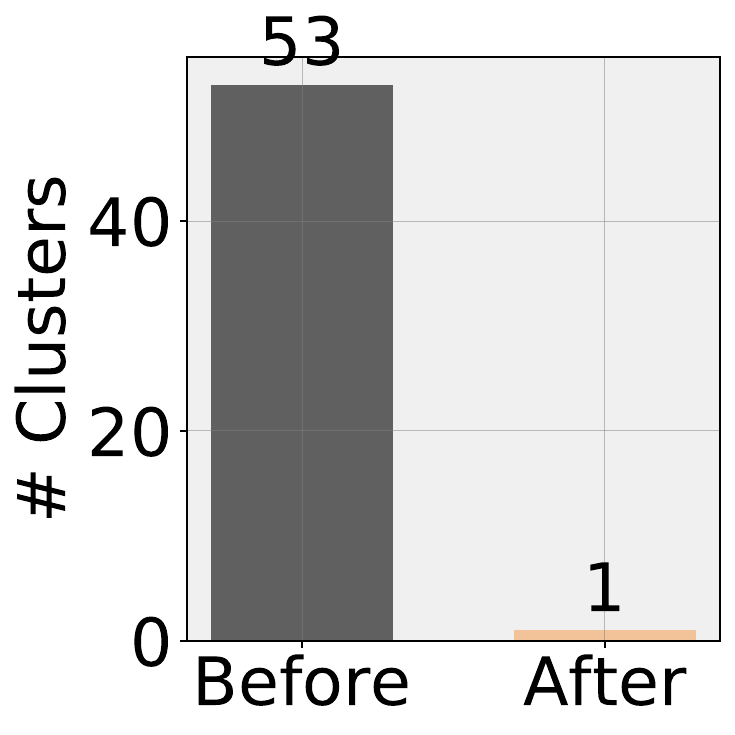}
    \centering
    \caption{Cluster Count Changes. Removing clusters that contain synthesis points leaves a single cluster, indicating the attacker's counterfeit object.}
    \centering
    \label{fig:3}
  \end{minipage}
\end{figure}

\vspace*{3mm}
We propose a cluster-based consistency measurement grounded in the fundamental concept that an object comprises more than one cluster and that point clusters attributed to a fake object lack consistency with clusters in historical frames.
In this section, we show the visualization results (Figure~\ref{fig:2}) and cluster count changes (Figure~\ref{fig:3}) when applied to a poisoned frame where an attacker injects a point set crafted to mimic a car.
\name~can identify fake objects in 97.2\% of cases as shown in D.CAR in Table~\ref{tab:eval_different_objects} in Sec.~\ref{sec:experiments}. Conversely, when our method is applied to benign frames, the absence of remaining clusters signifies the maintenance of object consistency across consecutive frames. 

\vspace*{-3mm}
\section{Optimal Thresholds for Anomaly Detection}
\vspace*{-3mm}
\label{supp_thresholdset}
\begin{figure}[htp]
\centering
% \vspace*{5mm} 
\includegraphics[width=0.9\linewidth]{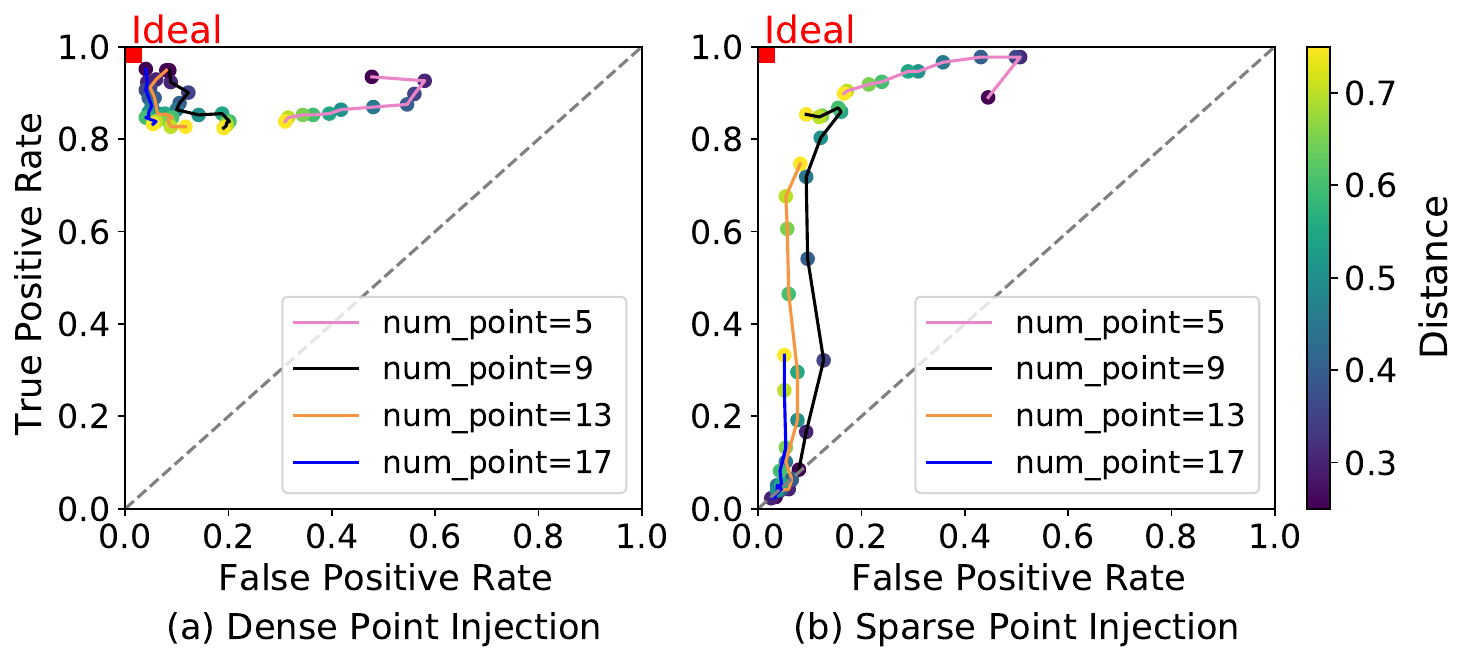}
\centering
\vspace*{2mm}
\caption{Determining Optimal Clustering Thresholds Based on False Positive (FPR) and True Positive Rates (TPR). 
Each colored line signifies a unique count threshold, while each colored dot represents a distinct distance threshold. Every red square marks the ideal scenario where the TPR is 1 and the FPR is 0. In determining the optimal threshold, proximity to the coordinate of the ideal case serves as a guiding factor.}
\vspace*{2mm}
\centering
\label{fig:1}
\end{figure}

% \vspace*{5mm} 
In determining the presence of anomalies, we apply DBSCAN, a prevalent spatial clustering technique, to all the points derived from merging the warped synthesis and the incoming frame. 
DBSCAN, by its nature, is a region-growing method that begins with an initial point and decides whether each subsequent point should be included in existing clusters or a new cluster. 
This decision-making process is significantly influenced by two threshold values: the minimum point requirement to form a cluster (count threshold) and the maximum distance between two points allowing them to belong in the same cluster (distance threshold).
Therefore, determining an optimal threshold set that considers both the false positive rate and the true positive rate is essential for detecting various attacks.

% However, finding the optimal threshold set is challenging due to the different sparsity presented by the attacks as illustrated in Figure~\ref{fig:1}.
% The figure reveals divergent trends for distance threshold between two attack types at a specific count threshold.
However, determining the ideal threshold set proves to be challenging due to the varied levels of sparsity exhibited by the attacks, as depicted in Figure~\ref{fig:1}. The figure underscores the diverging trends in distance thresholds between the two types of attacks at a specific count threshold.
For dense point injection, shorter distance thresholds are preferred, while sparse point injections benefit from longer distance thresholds.

We determine the optimal thresholds by considering the distance from the coordinate of the ideal case where TPR is 1 and FPR is 0. 
For dense point attacks, the optimal threshold is achieved with a count threshold of 17 and a distance threshold of 0.25. This threshold leads to an FPR of 4.5\% and a TPR of 95.2\% when a fake pedestrian is injected.
In the case of sparse point injection attacks, the optimal threshold is obtained with a count threshold of 9 and a distance threshold of 0.75, resulting in an FPR of 9.3\% and a TPR of 85.4\%.
% In the subsequent section, we describe the example of computing temporal consistency across consecutive frames using these determined optimal thresholds.

% \begin{wrapfigure}{r}{0.25\textwidth}
%     \vspace*{-7mm}     
%     \centering
%     \includegraphics[width=\linewidth,height=0.9\linewidth]{images/failure.pdf}
%     \vspace*{-8mm} 
%     \caption{Failure Case.}
%     \vspace*{-5mm} 
%     \label{fig:our_failure}
% \end{wrapfigure}

% \vspace{-3mm} 
\section{Failure Case Example}
\vspace{-3mm} 
\label{supp_failure}
\begin{figure}[h]
  \begin{minipage}[t]{0.24\linewidth}
    \centering
    \includegraphics[width=\linewidth,height=1\linewidth]{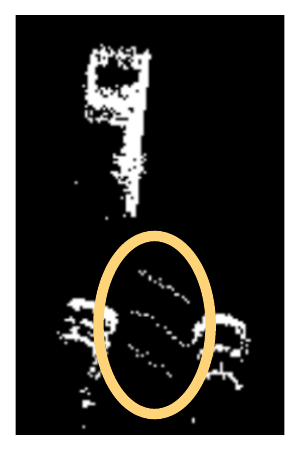}
  \end{minipage}
  \hfill
  \begin{minipage}[t]{0.75\linewidth}
  \vspace*{-20mm} 
    \caption{Representative Failure Case. The car marked with a yellow circle is a fake object created through a sparse injection attack. The perception module recognizes this fake car as part of the genuine vehicle on the right.}
    \vspace*{-15mm} 
    \label{fig:our_failure}
  \end{minipage}
\end{figure}

\vspace*{2mm}
Using spatial clustering for attack detection predominantly fails when spoofed points are near benign road objects (see Figure \ref{fig:our_failure}). While this leads to a false negative, it does not markedly influence driving decisions or result in the failure to trigger true alarms, considering the imperative to avoid collisions with the benign object that remains in place.

\end{document}